\documentclass{article}

\usepackage{PRIMEarxiv}
\usepackage{microtype}
\usepackage{graphicx}
\usepackage{subfigure}
\usepackage{booktabs} 
\usepackage{xcolor}
\usepackage{physics}
\usepackage{amsfonts}
\usepackage{amssymb}
\usepackage{bbm}
\usepackage{cite}
\usepackage[ruled]{algorithm2e}
\usepackage{bm}

\usepackage[T1]{fontenc}

\usepackage{longtable}

\usepackage{caption}
\usepackage{url}            
\usepackage{booktabs}       
\usepackage{amsfonts}       
\usepackage{nicefrac}       
\usepackage{microtype}      
\usepackage{lipsum}
\usepackage{fancyhdr}       
\usepackage{graphicx}       
\graphicspath{{media/}}     

\usepackage{academicons}
\definecolor{orcidlogocol}{HTML}{A6CE39}
 
\title{Fast Kernel Density Estimation with Density Matrices and Random Fourier Features
\thanks{\textit{\underline{Citation}}: 
\textbf{Joseph et al., Fast Kernel Density Estimation with Density Matrices and Random Fourier Features.}} 
}

\author{
  Joseph A. Gallego M., Juan F. Osorio, Fabio A. Gonz\'{a}lez \\
  MindLab \\
  Universidad Nacional de Colombia \\
  Bogot\'{a}, Colombia\\
  \texttt{\{jagallegom,josorior,fagonzalezo\}@unal.edu.co} 
}

\begin{document}

\maketitle

\begin{abstract}


Kernel density estimation (KDE) is one of the most widely used nonparametric density estimation methods. The fact that it is a memory-based method, i.e., it uses the entire training data set for prediction, makes it unsuitable for most current big data applications. Several strategies, such as tree-based or hashing-based estimators, have been proposed to improve the efficiency of the kernel density estimation method. The novel density kernel density estimation method (DMKDE) uses density matrices, a quantum mechanical formalism, and random Fourier features, an explicit kernel approximation, to produce density estimates. This method has its roots in the KDE and can be considered as an approximation method, without its memory-based restriction. In this paper, we systematically evaluate the novel DMKDE algorithm and compare it with other state-of-the-art fast procedures for approximating the kernel density estimation method on different synthetic data sets. Our experimental results show that DMKDE is on par with its competitors for computing density estimates and advantages are shown when performed on high-dimensional data. We have made all the code available as an open source software repository.


\end{abstract}

\section{Introduction}
\label{sec:introduction}
 In many applications we have a finite set of data and we would like to know what probability distribution has generated the data. From the point of view of statistical inference, this problem has played a central role in research and has inspired many methods that are based on the use of the density function. Also in machine learning there are many methods base on density estimation,  such as anomaly detection methods \cite{Lv2020}, generative models \cite{Liu2020}, agglomerative clustering \cite{Nakaya2010}, spatial analysis \cite{Borruso2008}, sequence-to-sequence models \cite{Peng2019}, among others.

Making certain assumptions on the probability model that generated the data leads to parametric estimation.  Another common approach is non-parametric estimation. The most representative non-parametric method is called Kernel Density Estimation (KDE) \cite{rosenblatt1956,parzen1962estimation} and can be understood as a weighted sum of density contributions that are centered on each data point. In this method, one has to choose a function called kernel and a smoothing parameter that controls the dispersion of the estimate. 
Given $n$ as the number of training data points, direct evaluation of the KDE on $m$ test data points requires $O(mn)$ kernel evaluations and $O(mn)$ additions and multiplications, i.e., if the number of training $n$ is sufficiently large, similar to the number of $m$ test points, its complexity is quadratic.
This makes it a very expensive process, especially for large data sets and higher dimensions as stated in \cite{gramacki2018nonparametric}. One approach to the problem of scalability of KDE focuses on finding a fast approximate kernel evaluation. According to \cite{Siminelakis2019}, we can identify three main lines of work: space partitioning methods, random sampling, and hashing-based estimators. In the Subsection \ref{subsec:kde_approximation_methods} we elaborate on each type of approximation. In \cite{gonzalez2021learning}, a novel KDE approach using random Fourier features and density matrices was proposed, which appears to be a promising way to scale up the KDE. This method uses an explicit Gaussian kernel approximation and a density matrix to produce density estimates. A more detailed explanation can be found in the subsection \ref{sec:quantum_kernel_density_estimation}. 

The goal of this paper is to compare, within a statistical experimental setup, several fast KDE implementations that are based on famous theoretical approaches for kernel density estimation approximation, including DMKDE, all the code is available as an open source software repository \cite{Gallego_M_Fast_Kernel_Density_2022}. The paper is organized as follows: Section 2 covers the background of random features, kernel density estimation; Section 3 presents some methods for kernel density estimation approximation; Section 4 presents the experimental setup and systematic evaluation, where seven synthetic data sets are used to compare DMKDE against other KDE approximation methods. This section shows that DMKDE outperformed other KDE approximation methods in terms of time consumed to make new predictions; finally, Section 5 discusses the conclusions of the paper and future research directions.

\section{Background and Related Work}
\label{sec:background}
\subsection{Kernel Density Estimation}\label{subsection:kde}
The multivariate kernel density estimator at a query point  $\bm{x}\in\mathbb{R}^d$ for a given random sample $X = \{\bm{x}_i\}_{i=1}^{n}$, where $n$ is the number of samples drawn from an unknown density $f$, is given by

\begin{equation}
\begin{aligned}
\hat{f}(\bm{x})&=\frac{1}{n} \sum_{i=1}^{n}|h|^{-1 / 2} k\left(h^{-1 / 2}\left(\bm{x}-\bm{x}_{i}\right)\right)
\end{aligned}
\label{eq:3}
\end{equation}
where $H$ is the $d\times d$ bandwidth matrix, which is positive definite and symmetric and $k:\mathbb{R}^d\rightarrow \mathbb{R}_{\geq 0}$ is the kernel function. Defining $k_{h}(\bm{u})=|h|^{-1 / 2} k\left(h^{-1 / 2} \bm{u} \right)$ we can write \ref{eq:3} as

\begin{equation}
\hat{f}(\bm{x})=\frac{1}{n} \sum_{i=1}^{n} k_h\left(\bm{x}-\bm{x}_{i}\right)
\label{eq:4}
\end{equation}

Throughout the paper we will work with the \textit{rescaled} multivariate KDE version written in the Equation \ref{eq:4}. We will also assume that $h = \sigma^2\mathbf{I}_{d}$ where $\sigma^2$ represents the bandwidth in each dimension, $\mathbf{I}_{d}$ is the $d \times d$ identity matrix, and the kernel will be Gaussian expressed by

\begin{equation}
K_{\gamma}(\bm{x}, \bm{y})=(2 \pi)^{-d / 2} \exp \left(-\gamma ||\bm{x}-\bm{y}||^2\right).
\label{eq:5}
\end{equation}
where $\gamma=1/(2\sigma^2)$.

All of these assumptions lead to the functional form

\begin{equation}
\hat{f}_{\gamma, X}(\bm{x})=\frac{1}{n(\pi / \gamma)^{\frac{d}{2}}} \sum_{i=1}^{n} e^{-\gamma\left\|\bm{x}-\bm{x}_{i}\right\|^{2}}
\label{eq:6}
\end{equation}
of the KDE estimator for a query point $x\in\mathbb{R}^d$ where we define $\gamma = \frac{1}{2\sigma}$.

Kernel density estimation has multiple applications some examples include: to visualize clusters of crime areas using it as a spatio-temporal modification \cite{Nakaya2010}; to assess injury-related traffic accidents in London, UK \cite{Anderson2009}, to generate the intensity surface in a spatial environment for moving objects couple with time geography \cite{Downs2010}; to make a spatial analysis of city noise with information collected from a French mobile application installed on citizens' smartphones using their GPS location data \cite{GuardnacciaClaudio2021}; to generate new samples from a given dataset by dealing with unbalanced datasets \cite{Kamalov2020}; to process the image by subtracting the background in a pixel-based method \cite{Lee2012}; to propose an end-to-end pipeline for classification \cite{Kristan2011}; to propose a classification method using density estimation and random Fourier features \cite{gonzalez2021learning}.


\subsection{Random Fourier Features}\label{subsect:random_fourier_feature}
Using random Fourier features (RFF) to approximate kernels was initially presented in \cite{rahimi2007rff}. In this method, the authors approximate a shift invariant kernel by an inner product in an explicit Hilbert spaces. Formally, given a shift invariant kernel $k: \mathbb{R}^d\times\mathbb{R}^d \rightarrow \mathbb{R}$ they build a map $\phi_{\text{rff}} : \mathbb{R}^d \rightarrow \mathbb{R}^D$ such that

\begin{equation}
k(x,y) \approx \phi_{\text{rff}}(\bm{x})^* \phi_{\text{rff}}(\bm{y})
\label{eq:7}
\end{equation}
for all $\{\bm{x},\bm{y}\}\subseteq \mathbb{R}^d$, and in the case of the Gaussian kernel the mapping is defined as follows  
\begin{equation}
\phi_{\text{rff}}(\bm{x}) = \sqrt{2}\cos(\bm{\omega}^*\bm{x} + b)
\label{eq:rff_gaussian}
\end{equation}
where $\bm{w}$ is sampled from $\mathcal{N}(\bm{0}, \bm{I}_D)$ and $b$ is samples from Uniform(0,2$\pi$). The main theoretical result supporting \ref{eq:7} comes from an instance of Bochner's theorem \cite{rudin1962fourier} which states that a shift invariant continuous kernel is the Fourier transform of a nonnegative probability measure. This methodology suggests that we can compute shift-invariant kernel approximations via a sampling strategy.

\subsection{Quantum Kernel Density Estimation Approximation}
\label{sec:quantum_kernel_density_estimation}
The central idea of the method density matrix kernel density estimation (DMKDE), introduced by \cite{gonzalez2021learning} and systematically evaluated in \cite{josephquantum}, is to use density matrices along with random Fourier features to represent arbitrary probability distributions by addressing the important question of how to encode probability density functions in $\mathbb{R}^n$ into density matrices. The overall process is divided into a training and a testing phases. The training phase is defined as follows:

\begin{enumerate}
    \item Input: a set of $n$ $d$-dimensional samples $\bm{x}_1,\cdots, \bm{x}_n$, $D$ number of random Fourier features and a bandwidth parameter $\gamma \in \mathbb{R}$
    \item Sampling of the vector  $\bm{w}$ and b as explained in Subsection \ref{subsect:random_fourier_feature}
    \item Mapping: compute the random Fourier feature vector for each data point as explained in Equation \ref{eq:rff_gaussian}. 
    \item $\bm{x}_i$ as $\bm{z}_i = \phi_{\text{rff}}(\bm{x}_i)$
    \item Compute a density matrix as $\rho= \frac{1}{n}\sum_{i=1}^n \bm{z}_i\bm{z}_i^T$
\end{enumerate}

The testing phase is as follows:

\begin{enumerate}
    \item Apply step $3$ to each testing data point.
    \item The density estimation of a testing point $\bm{x}$ is calculate using the Born's rule
    $$\hat{f}_{\rho}(\bm{x})= \frac{\phi_{\text{rff}}(\bm{x})^T\rho\phi_{\text{rff}}(\bm{x})}{\mathcal{Z}}$$
    where the normalizing constant is: $\mathcal{Z}=\left(\pi/ (2\gamma)\right)^{d/2}$
\end{enumerate}

Note that this algorithm does not need to store each training data point, but a square matrix whose dimensions are equal to $D \times D$ suffices computed in step 5. Moreover, the time consumed in estimating a new data point does not depend on the training size. The $\rho$-density matrix is one of the main building blocks used in quantum physics to capture classical and quantum probability in a given physical system.  This formalism was conceived by \cite{von1927wahrscheinlichkeitstheoretischer} as the foundation of quantum statistical mechanics. The density matrix describes the states of the quantum system and explains the relationship between the pure state and the mixed states of the system. Define the quantum state of a system in pure state as $\psi$. Then the $\rho$-density matrix  for the pure state $\psi$ is defined as $\rho:= |\psi\rangle \langle \psi|$. Let us now consider a quantum ensemble system of $n$ systems (objects) that are not in the same state. Let $p_i:=n_i/n$ where $n_i$ is the number of systems that are in state $|\psi_i\rangle$ and $\sum n_i = n$ . Therefore, the mixed density matrix is defined as $\rho_{mix} = \sum_i p_i \rho_i^{pure}=\sum_i |\psi_i\rangle \langle \psi_i|$.

A matrix factorization of the $\rho$-matrix  simplifies the computation of the matrix as $\rho=V^*\Lambda V$, where $V\in \mathbb{R}^{r\times D}, \Lambda \in \mathbb{R}^{r\times r}$ is a diagonal matrix and $r < D$ is the reduced rank of the factorization. Thus, using this matrix factorization, the method called DMKDE-SGD is expressed as: $\hat{f}_{\rho}= \frac{1}{\mathcal{Z}} || \Lambda ^{1/2} V \phi_{\text{rff}}(\bm{x})||^2$. This method reduces the time required for a new density to $O(Dr)$.

\section{KDE Approximation Methods} \label{subsec:kde_approximation_methods}
Fast density estimation methods are of paramount importance in several applications, as shown above in \ref{subsection:kde}. There are four main approaches for fast kernel density estimation approximation. First, space partitioning methods do not work well in high dimensions and make use of geometric data structures to partition the space, and achieve speedup by limiting the contribution of data points to the kernel density using partitions. Second, random sampling focuses on randomly sampling the kernel density and gives good results in high dimensions. Third, the most recent methods and the most promising state-of-the-art tools are hash-based estimators for kernel density estimation, in which a hash structure is used to construct close distance boxes that allow the calculation of few distances in the prediction step \cite{Backurs2019}.  Here, a sampling scheme is used, where points are sorted into buckets thanks to a hash function whose main objective is to send similar objects to the same hash value. Finally, the novel DMKDE as explained above in \ref{sec:quantum_kernel_density_estimation} can approximate kernel density estimation. The following algorithms including DMKDE are used in the experimental setup.
\begin{itemize}
    \item \textit{Tree kernel density estimation (TREEKDE)}: the density approximation is obtained using the partitioning of the space through recursive segmentations of the space into smaller sections. By obtaining the partition, we can approximate the kernel of a specific point by traversing the tree, without having to explicitly evaluate the kernel function \cite{maneewongvatana1999s}.
    \item \textit{Kernel Density Estimation using k-dimensional Tree (KDEKDT)}: this method addresses the k-nearest neighbor problem using a kd-tree structure that generalizes two-dimensional Quad-trees and three-dimensional Oct-trees to an arbitrary number of dimensions. Internally, it is a binary search tree that partitions the data into nested orthotopic regions that are used to approximate the kernel at a specific point \cite{bentley1975multidimensional}.
    \item \textit{Kernel Density using Ball Tree (KDEBT)}: in this method the Ball Tree structure is used to avoid the inefficiencies of KD trees in high dimensionality spaces. In these spaces, the Ball Tree divides the space into nested hyperspheres \cite{omohundro1989five}.
\end{itemize}

\section{Experimental Evaluation}
\label{sec:experimental_evaluation}

In this section, we systematically assess the performance of DMKDE on various synthetic data sets and compare it with kernel density estimation approximation methods.

\subsubsection{Data sets and experimental setup}

We used seven synthetic data sets to evaluate DMKDE against kernel density approximation methods. The data sets are characterized as follows:

\begin{itemize}
    \item The data set \textit{Arc} corresponds to a two-dimensional random sample drawn from a random vector $\bm{X}=(X_1,X_2)$ with probability density function given by $$f(x_1,x_2)=\mathcal{N}(x_2|0,4)\mathcal{N}(x_1|0.25x_2^2,1)$$
              where $\mathcal{N}(u|\mu,\sigma^2)$ denotes the density function of a normal distribution with mean $\mu$ and variance $\sigma^2$. \cite{Papamakarios2017} used this data set to evaluate his neural density estimation methods.
    \item The data set \textit{Potential 1} corresponds to a two-dimensional random sample drawn from a random vector $\bm{X}=(X_1,X_2)$ with probability density function given by \begin{scriptsize}
    $$f(x_1,x_2)=\frac{1}{2}\left(\frac{||\bm{x}||-2}{0.4}\right)^2  - \ln{\left(\exp\left\{-\frac{1}{2}\left[\frac{x_1-2}{0.6}\right]^2\right\}+\exp\left\{-\frac{1}{2}\left[\frac{x_1+2}{0.6}\right]^2\right\}\right)}$$
    \end{scriptsize}
    with a normalizing constant of approximately 6.52 calculated by Monte Carlo integration.
    \item The data set \textit{Potential 2} corresponds to a two-dimensional random sample drawn from a random vector $\bm{X}=(X_1,X_2)$ with probability density function given by  $$f(x_1,x_2)=\frac{1}{2}\left[ \frac{x_2-w_1(\bm{x})}{0.4}\right]^2$$ 
  where $w_1(\bm{x})=\sin{(\frac{2\pi x_1}{4})}$ with a normalizing constant of approximately 8 calculated by Monte Carlo integration.
    \item The data set \textit{Potential 3} corresponds to a two-dimensional random sample drawn from a random vector $\bm{X}=(X_1,X_2)$ with probability density function given by \begin{scriptsize}$$f(x_1,x_2)=  - \ln{\left(\exp\left\{-\frac{1}{2}\left[\frac{x_2-w_1(\bm{x})}{0.35}\right]^2\right\}+\exp\left\{-\frac{1}{2}\left[\frac{x_2-w_1(\bm{x})+w_2(\bm{x})}{0.35}^2\right]\right\}\right)}$$\end{scriptsize}
     where $w_1(\bm{x})=\sin{(\frac{2\pi x_1}{4})}$ and $w_2(\bm{x})=3 \exp \left\{-\frac{1}{2}\left[ \frac{x_1-1}{0.6}\right]^2\right\}$ with a normalizing constant of approximately 13.9 calculated by Monte Carlo integration.
    \item The data set \textit{Potential 4} corresponds to a two-dimensional random sample drawn from a random vector $\bm{X}=(X_1,X_2)$ with probability density function given by \begin{scriptsize}$$f(x_1,x_2)=  - \ln{\left(\exp\left\{-\frac{1}{2}\left[\frac{x_2-w_1(\bm{x})}{0.4}\right]^2\right\}+\exp\left\{-\frac{1}{2}\left[\frac{x_2-w_1(\bm{x})+w_3(\bm{x})}{0.35}^2\right]\right\}\right)}$$\end{scriptsize}
     where $w_1(\bm{x})=\sin{(\frac{2\pi x_1}{4})}$, $w_3(\bm{x})=3 \sigma \left(\left[ \frac{x_1-1}{0.3}\right]^2\right)$, and $\sigma(x)= \frac{1}{1+\exp(x)}$ with a normalizing constant of approximately 13.9 calculated by Monte Carlo integration.
    \item The data set \textit{2D mixture} corresponds to a two-dimensional random sample drawn from the random vector $\bm{X}=(X_1, X_2)$ with a probability density function given by $$f(\bm{x}) = \frac{1}{2}\mathcal{N}(\bm{x}|\bm{\mu}_1,\bm{\Sigma_1}) + \frac{1}{2}\mathcal{N}(\bm{x}|\bm{\mu}_2,\bm{\Sigma_2})$$
     with means and covariance matrices $\bm{\mu}_1 = [1, -1]^T$, $\bm{\mu}_2 = [-2, 2]^T$, $\bm{\Sigma}_1=\left[\begin{array}{cc} 1 & 0 \\ 0 & 2 \end{array}\right]$, and $\bm{\Sigma}_1=\left[\begin{array}{cc} 2 & 0 \\ 0 & 1 \end{array}\right]$
    \item The data set \textit{10D-mixture} corresponds to a 10-dimensional random sample drawn from the random vector $\bm{X}=(X_1,\cdots,X_{10})$ with a mixture of four diagonal normal probability density functions $\mathcal{N}(X_i|\mu_i, \sigma_i)$, where each $\mu_i$ is drawn uniformly in the interval $[-0.5,0.5]$, and the $\sigma_i$ is drawn uniformly in the interval $[-0.01, 0.5]$. Each diagonal normal probability density has the same probability of being drawn $1/4$.
\end{itemize}

The functions from \textit{Potential 1} to \textit{4} were presented in \cite{rezende} to test their normalizing flow algorithms. The \textit{ARC} data set was presented in \cite{Papamakarios2017} to test his autoregressive models.

\begin{figure*}[t]
\begin{centering}
\includegraphics[scale=0.37]{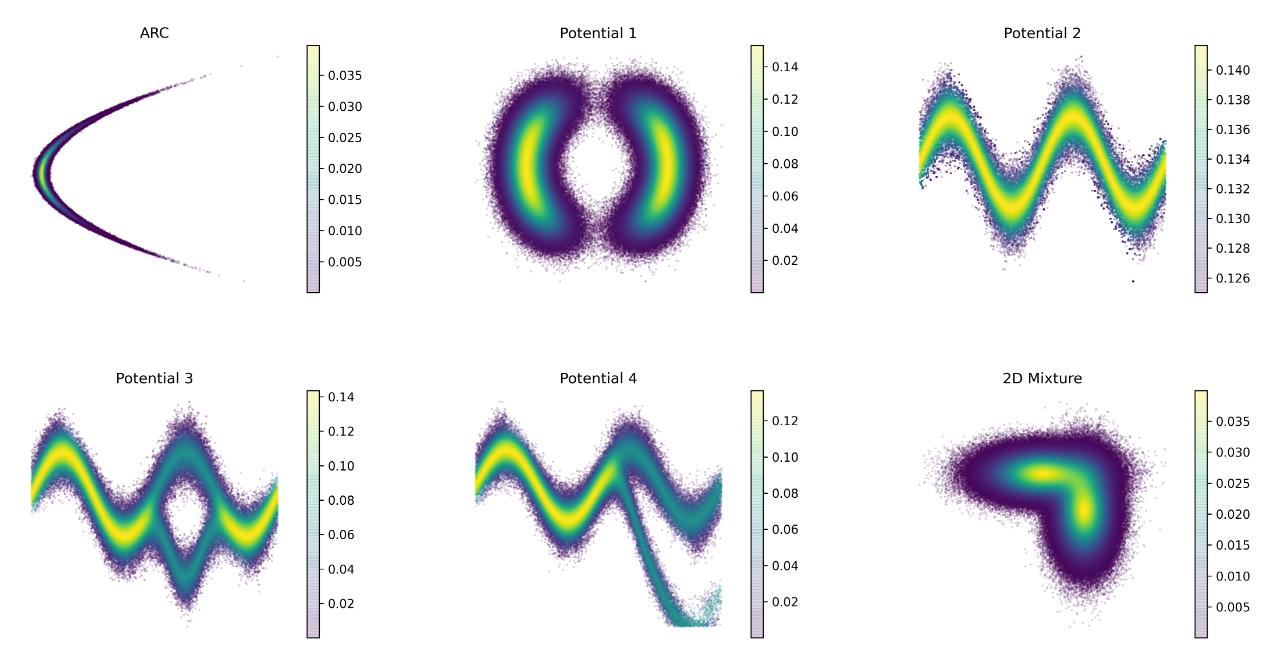}
\par\end{centering}
\caption{True density of each data set: arc, potential 1 to 4, and 2D mixture. High density points are colored as yellow and low-density points are colored as white. \label{fig:model}}
\vskip -0.33in
\end{figure*}

For this experiment, we compared DMKDE with different methods of approximate kernel density estimation. We used for this comparison: (1) a raw implementation of kernel density estimation using Numpy (RAWKDKE, (2) a naive implementation of kernel density estimation using the KDE.py library (NAIVEKDE) \footnote{Implementation of KDE.py: \url{https://github.com/Daniel-B-Smith/KDE-for-SciPy/blob/master/kde.py}}, (3) tree-based kernel density estimation (TREEKDE), (4) kernel density estimation using a k-dimensional tree (KDEKDT), and (5) kernel density using a ball tree (KDEBT). Details of NAIVEKDE, KDEKDT, and KDEBT can be found above in \ref{subsec:kde_approximation_methods}. In this experimental setup, we did not compare with the hashing-based approach due to the restriction of the algorithm implementations. For all experiments, the Gaussian kernel was used. Two different types of experiments were performed. In the first, we evaluated the accuracy of each of the algorithms on each data set. In the second, we evaluated the time it takes to make a new density prediction on the training set. Each run was performed with different training set sizes, using a scale of $10^i$ where $i \in \{1,2,3,4,5\}$, and we set the test set to $10^4$ test examples. The spread parameter was found using a 5-fold cross validation for each data set and each training size in a logarithmic scale $\gamma \in \{2^{-20},\cdots,2^{20}\}$. The optimal number of random Fourier features used by DMKDE-SGD was searched in the set $\{50,100,500,1000\}$. After finding the best-hyperparameter, several attempts were performed for each algorithm in each data set.

We measure the efficacy of each algorithm on each data set using the $L_1$-error also known as \textit{average error}. Some advantages over $L_2$-error are outlined in \cite{devroye1985nonparametric}. The $L_1$-error is defined over $n$ samples by the following equation: $MAE = \frac{1}{n} \sum_{i=1}^n |\hat{f}(x) - f(x)|.$ In \cite{devroye1985nonparametric}, the author shows that the loss $L_1$ loss $\int |\hat{f}-f| $ is invariant under monotone transformations of the coordinate axes and points out that it is related to the maximum error made if we were estimating the probabilities of all Borel sets of $\hat{f}$ and $f$ respectively. Efficiency was evaluated using CPU time in milliseconds (ms), which defines the amount of time it takes the central processing unit (CPU) to execute its processing instructions to compute the evaluation query. We used the built-in \textbf{time} package in the Python programming language to measure the elapsed time in prediction time for each algorithm.

\subsubsection{Results and discussion}

\begin{table*}[tbh]
\vskip -0.3in
\caption{Efficacy test results measured in MAE for training size $1 \times 10^5$}
\label{Table:mae}
\vskip -0.15in
\begin{center}
\begin{sc}
\begin{tiny}
\begin{tabular}{lccccccr}
\toprule
\toprule
DATA SET & RAW & NAIVE & TREE & KDBTREE & KDKDTREE & DMKDE-SGD \\

       \toprule
        ARC & $\bm{0.0012}\pm$2E-4 &  $\bm{0.0012}\pm$1E-4 &  $\bm{0.0012}\pm$2E-4 &  $\bm{0.0012}\pm$1E-4 &  $\bm{0.0012}\pm$1E-4 & $0.0080\pm$1E-4 \\
        2D MIXTURE &  $\bm{0.0010}\pm$1E-4 &  $\bm{0.0010}\pm$1E-4 &  $\bm{0.0010}\pm$3E-4 &  $\bm{0.0010}\pm$5E-4 &  $\bm{0.0010}\pm$5E-4 & $0.0016\pm$2E-4 \\
        10D MIXTURE & 2.5282$\pm$3E-4 & 2.5282$\pm$3E-4 & 583000$\pm$0.0004 & 2.6216$\pm$2E-4 & 2.5282$\pm$3E-4 &  $\bm{1.7420}\pm$1E-4 \\
        POTENTIAL 1 & 0.0046$\pm$1E-4 & 0.0046$\pm$1E-4 & 0.0046$\pm$3E-4 & 0.0046$\pm$2E-4 & 0.0046$\pm$6E-4 &  $\bm{0.00334}\pm$3E-4 \\
        POTENTIAL 2 & 0.0456$\pm$2E-4 & 0.0456$\pm$2E-4 & 0.0456$\pm$0.0004 & 0.0456$\pm$0.0007 & 0.0456$\pm$0.0007 &  $\bm{0.0332}\pm$2E-4 \\
        POTENTIAL 3 &  $\bm{0.0182}\pm$1E-4 &  $\bm{0.0182}\pm$1E-4 &  $\bm{0.0182}\pm$5E-4 & 0.0182$\pm$6E-4 & 0.0182$\pm$6E-4 & 0.0299$\pm$2E-4 \\
        POTENTIAL 4 & $\bm{0.0190}\pm$2E-4 & $\bm{0.0190}\pm$1E-4 & $\bm{0.0190}\pm$2E-4 & $\bm{0.0190}\pm$5E-4 & $\bm{0.0190}\pm$8E-4 & 0.0235$\pm$3E-4 \\
\bottomrule
\end{tabular}
\end{tiny}
\end{sc}
\end{center}
\vskip -0.3in
\end{table*}

\begin{table*}[tbh]
\caption{Efficiency test results in millisecond (ms) for training size $1 \times 10^5$}
\label{Table:time}
\vskip -0.5in
\begin{center}
\begin{sc}
\begin{scriptsize}
\begin{tabular}{lccccccr}
\toprule
\toprule
       DATA SET & RAW & NAIVE & TREE & KDBTREE & KDKDTREE & DMKDE-SGD \\
       \toprule
        ARC & 52400$\pm$500 & 103000$\pm$447 & 24700$\pm$171 & 49200$\pm$700 & 56000$\pm$141 & $\bm{4330\pm145}$ \\
        2D MIXTURE & 42400$\pm$282 & 68000$\pm$707 & 35900$\pm$282 & 54800$\pm$505 & 62000$\pm$577 & $\bm{3500\pm190}$  \\
        10D MIXTURE & 88000$\pm$115 & 123000$\pm$1527 & 583000$\pm$435 & 190000$\pm$436 & 193000$\pm$577 & $\bm{504\pm110}$  \\
        POTENTIAL  1 & 46100$\pm$378 & 76000$\pm$547 & 22100$\pm$141 & 39901$\pm$212 & 52400$\pm$141 & $\bm{3640\pm311}$  \\
        POTENTIAL 2 & 28800$\pm$212 & 70000$\pm$0.000 & 4920$\pm$210 & 29400$\pm$302 & 36700$\pm$700 & $\bm{3630\pm106}$  \\
        POTENTIAL 3 & 30200$\pm$0.000 & 72000$\pm$0.000 & 19900$\pm$410 & 41100$\pm$520 & 48900$\pm$0.000 & $\bm{3280\pm176}$  \\
        POTENTIAL 4 & 30700$\pm$0.000 & 68000$\pm$707 & 54700$\pm$424 & 54800$\pm$0.000 & 64000$\pm$0.000 & $\bm{3700\pm158}$  \\
\bottomrule
\end{tabular}
\end{scriptsize}
\end{sc}
\end{center}
\vskip -0.2in
\end{table*}

\begin{figure*}[tbh]
\begin{centering}
\includegraphics[scale=0.5]{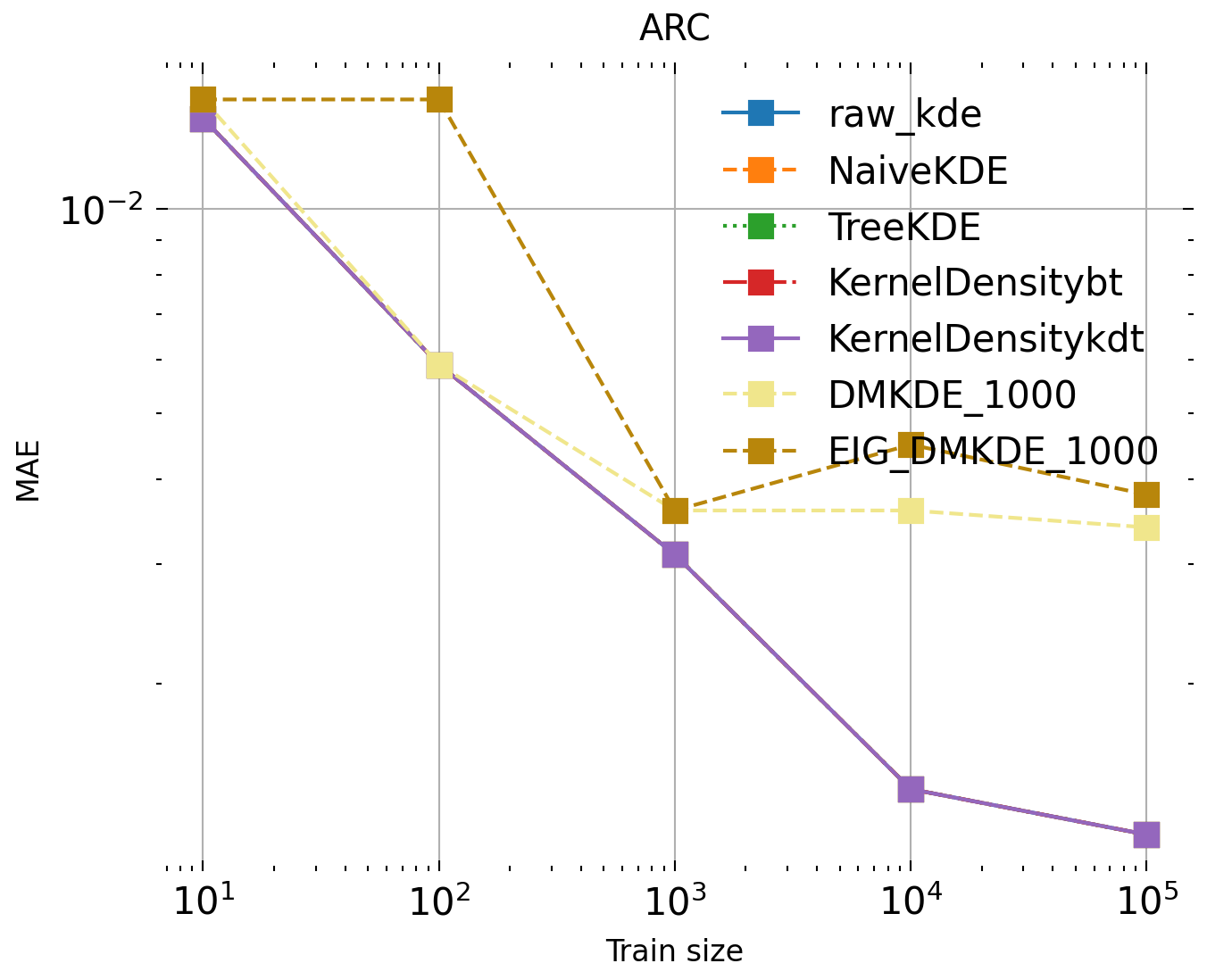}
\includegraphics[scale=0.5]{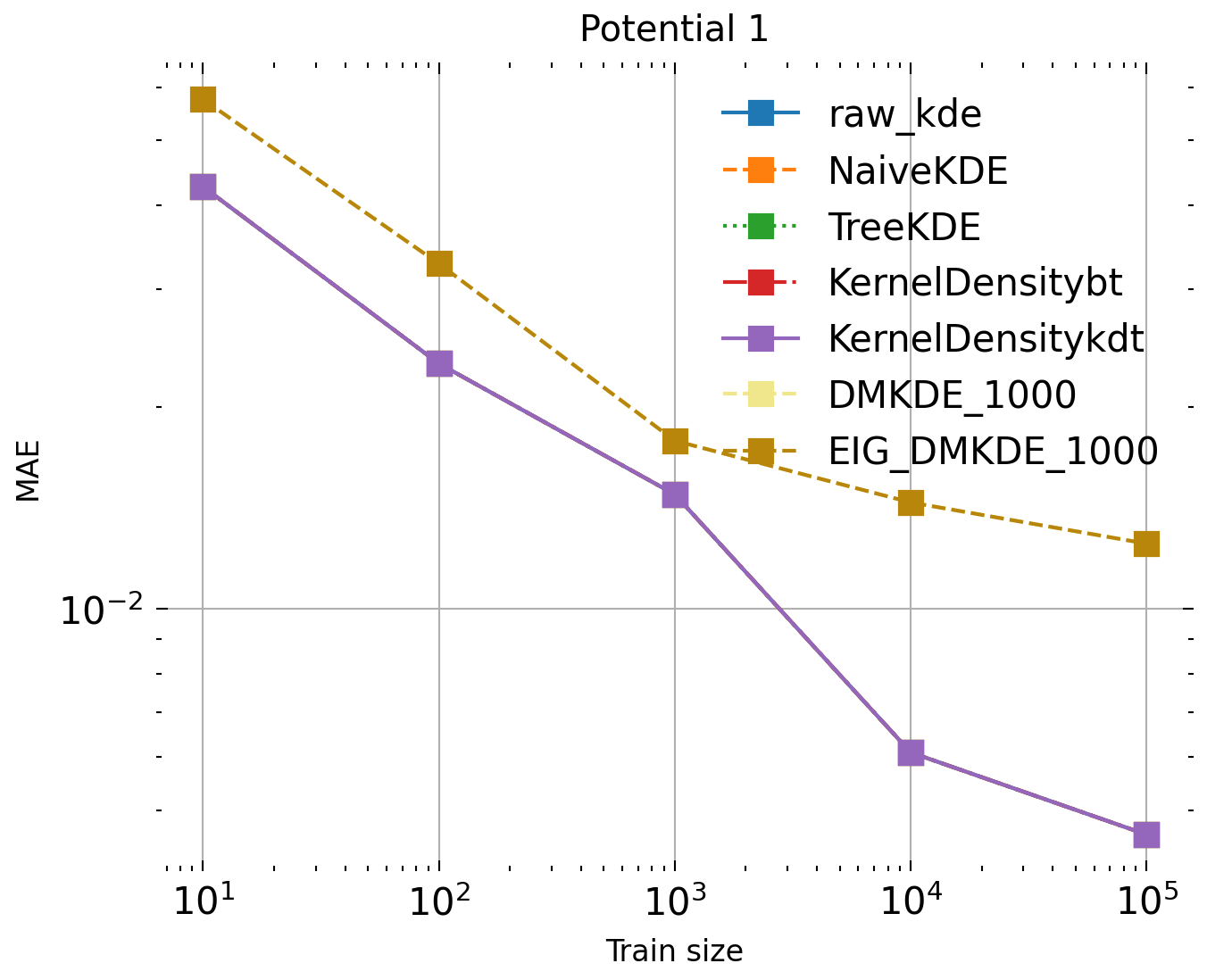}
\includegraphics[scale=0.5]{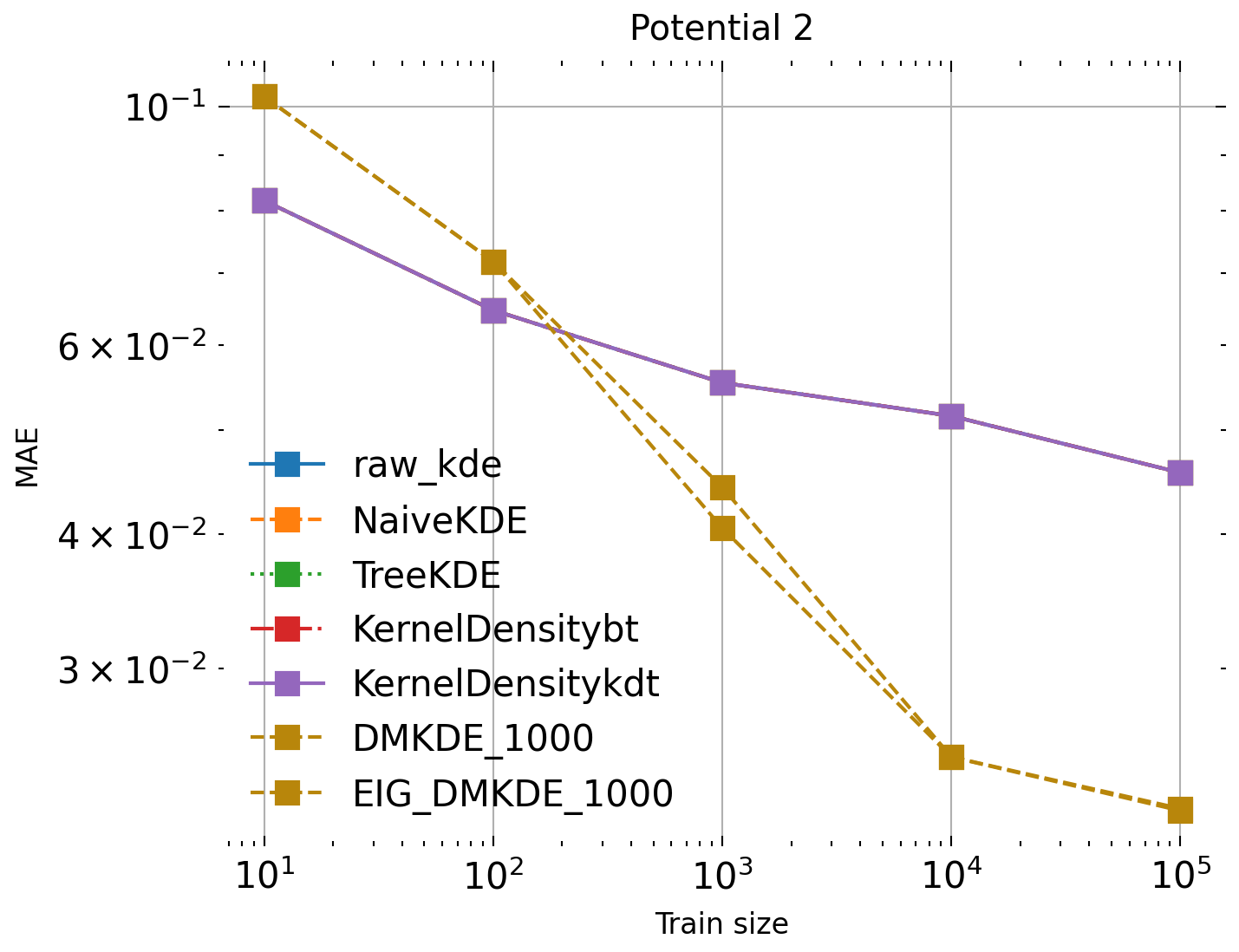}
\includegraphics[scale=0.5]{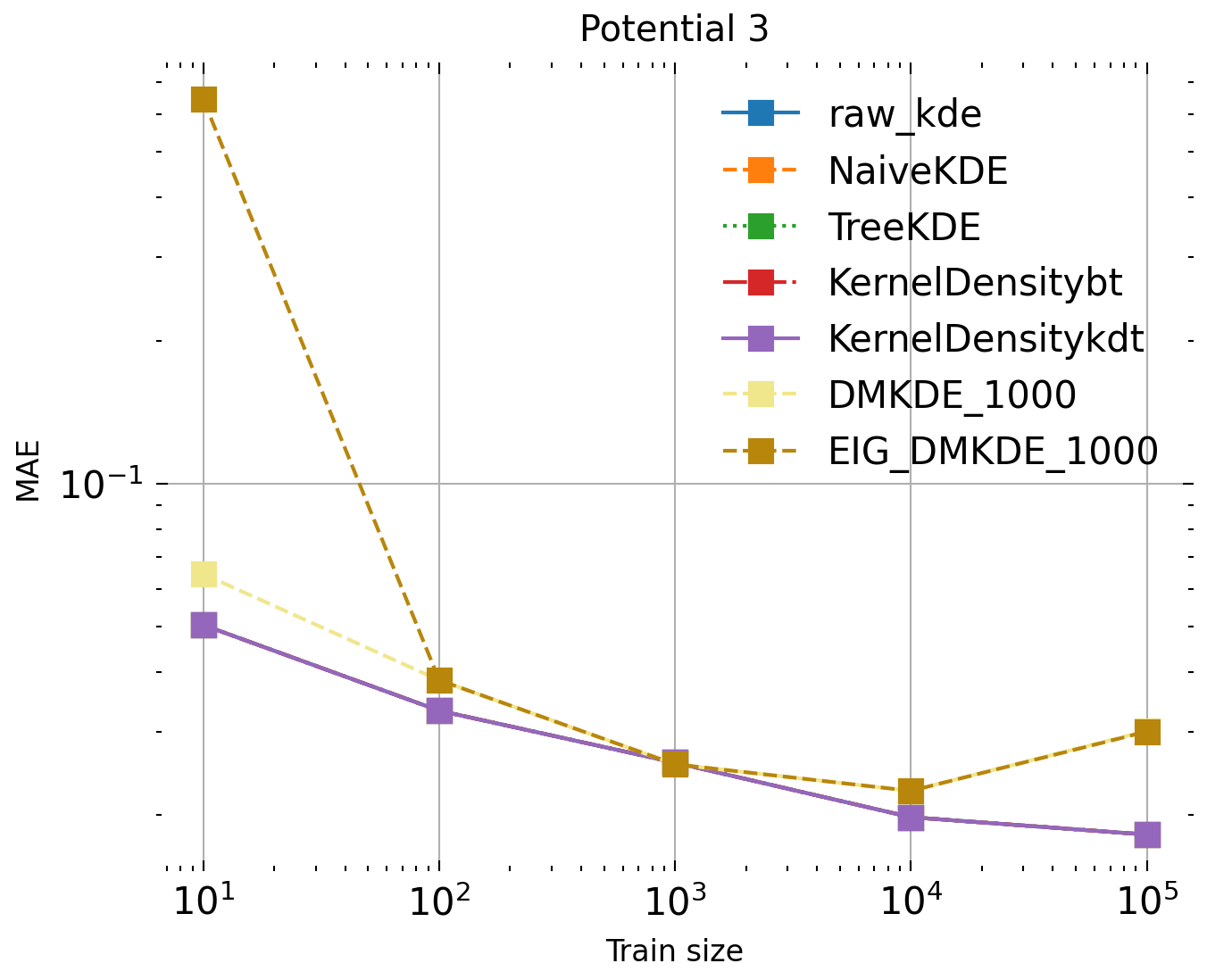}
\includegraphics[scale=0.5]{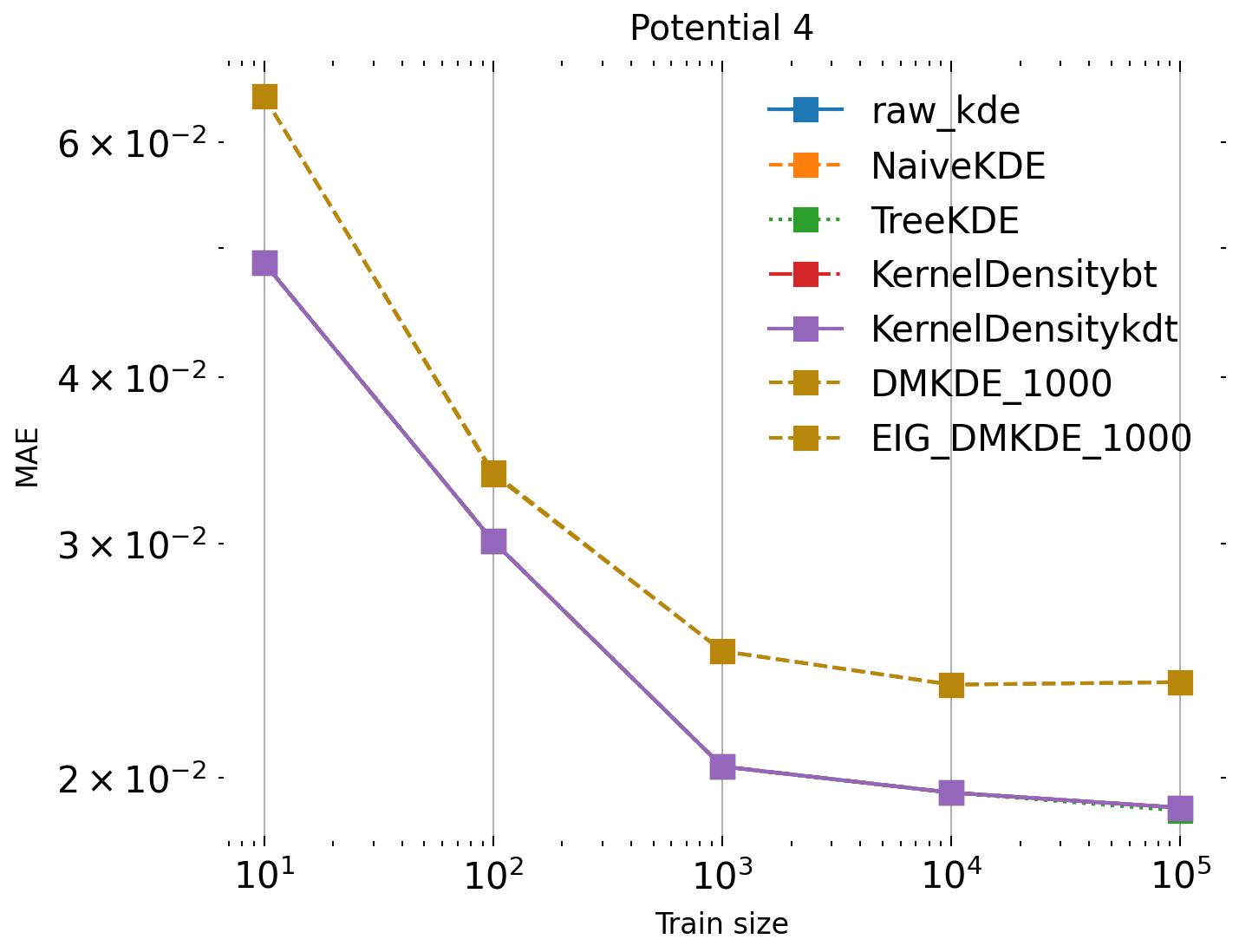}
\includegraphics[scale=0.5]{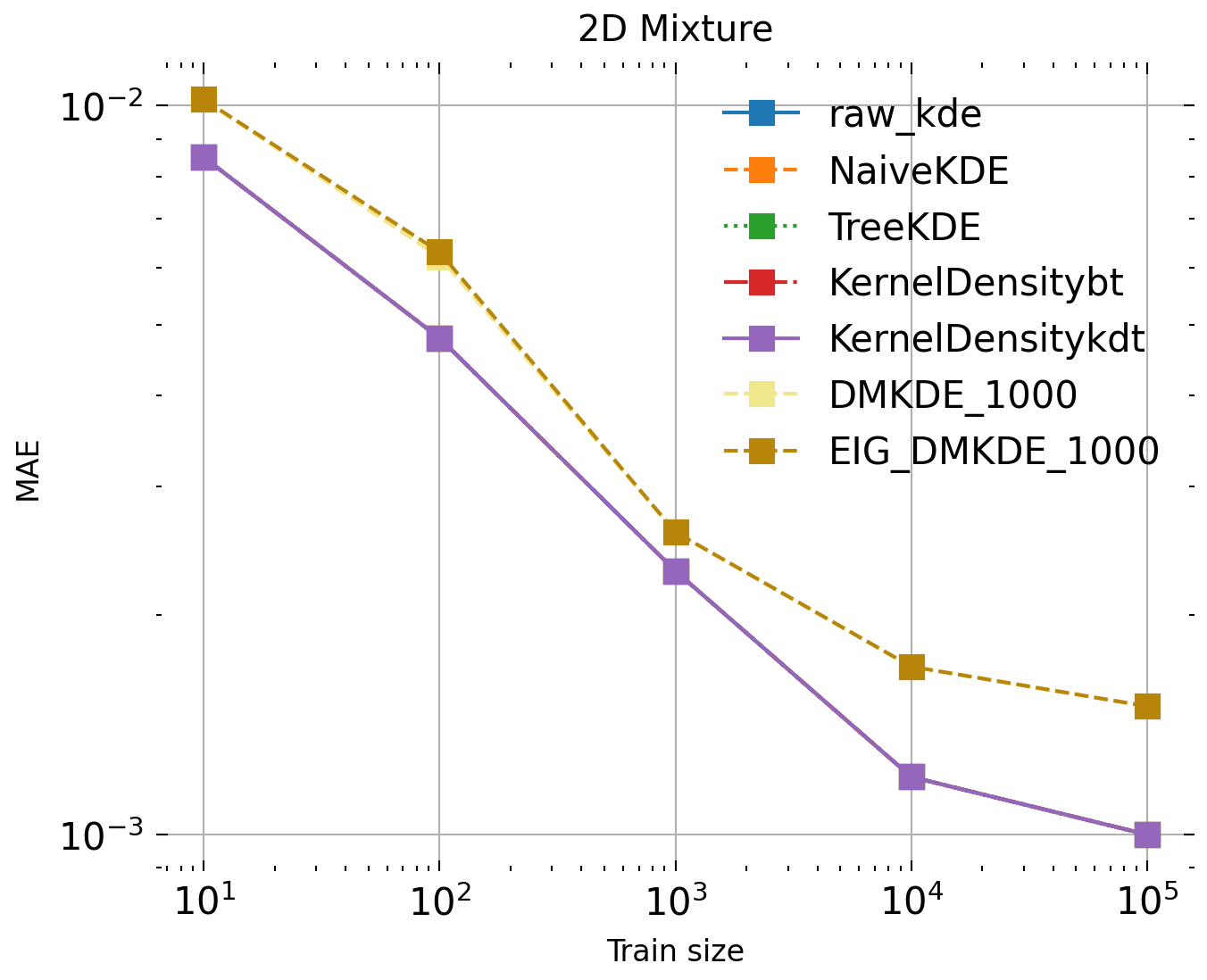}
\includegraphics[scale=0.5]{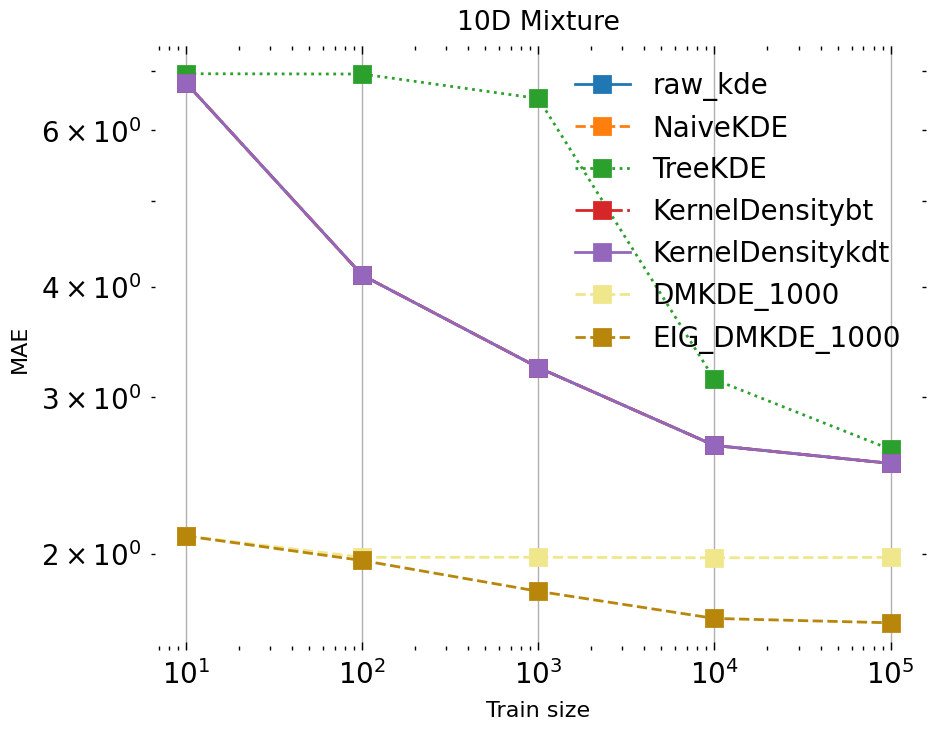}
\par\end{centering}
\caption{Comparison of the efficacy of each algorithm on six synthetic data sets. The x-axis is a logarithmic scale of $10^i$ where $i \in \{1,\cdots,5\}$. The y-axis represents the mean average error between  the prediction of the algorithm and the true density.}\label{fig:comparison_efficacy}
\end{figure*}

\begin{figure*}[tbh]
\begin{centering}
\includegraphics[scale=0.5]{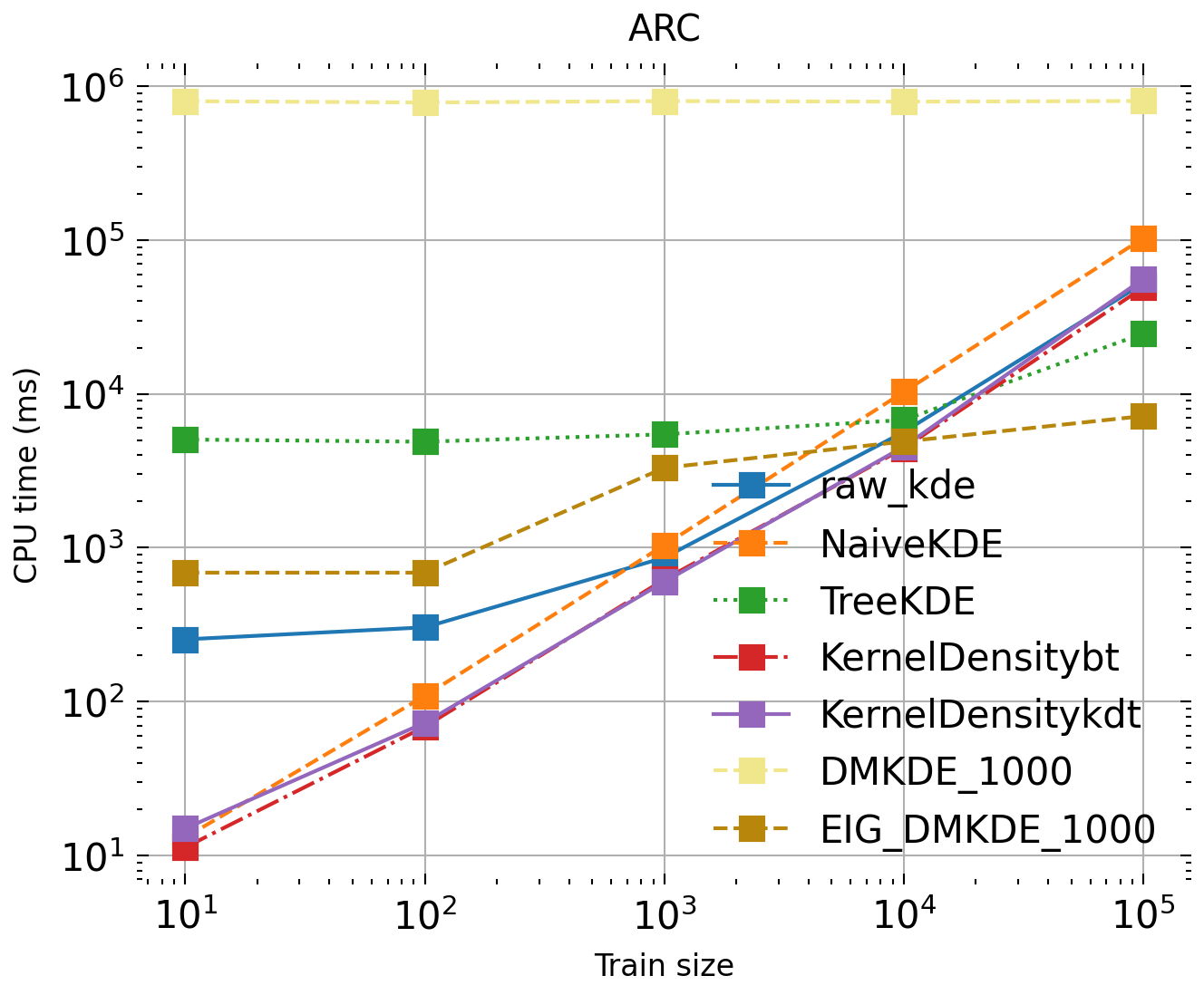}
\includegraphics[scale=0.5]{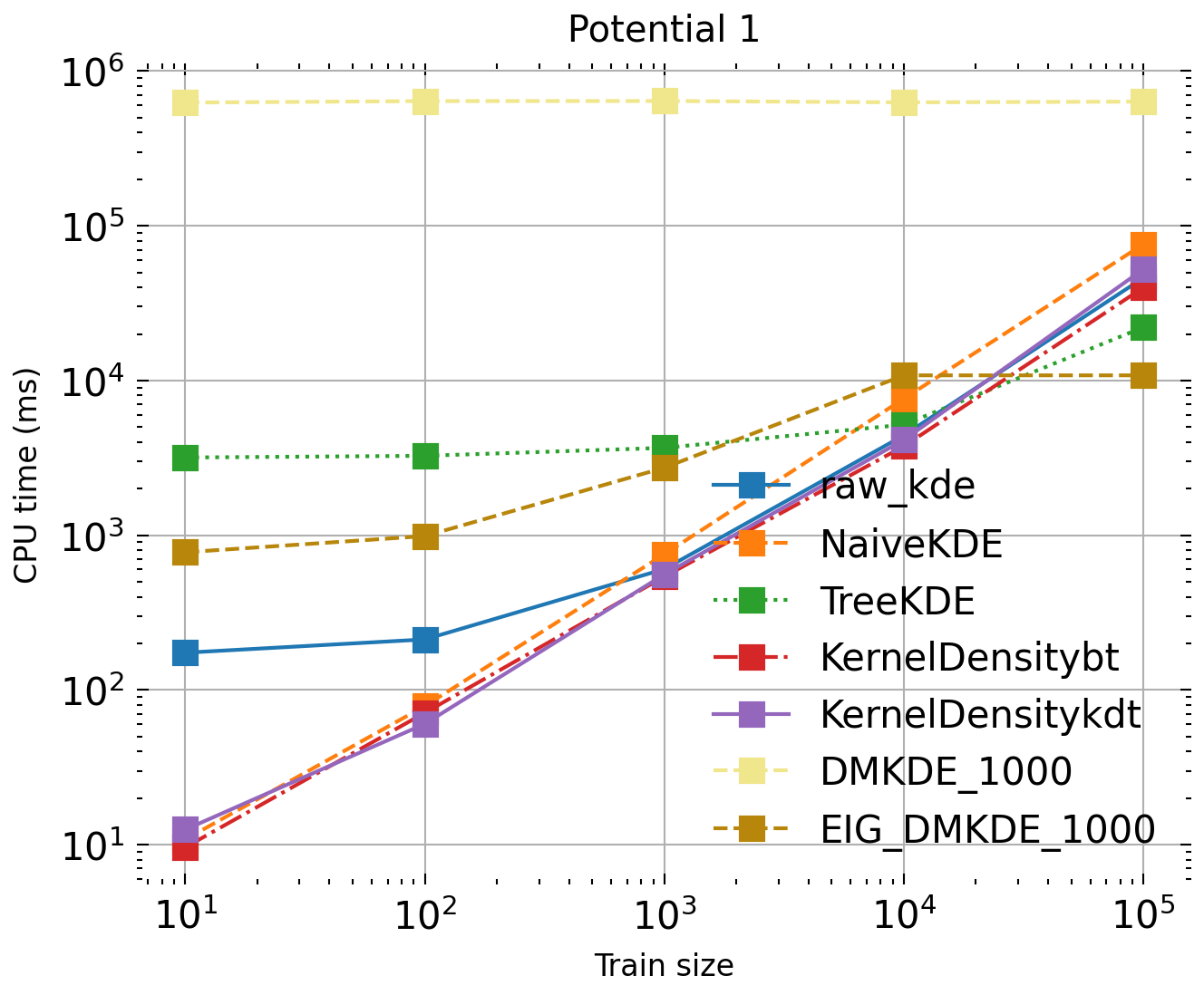}
\includegraphics[scale=0.5]{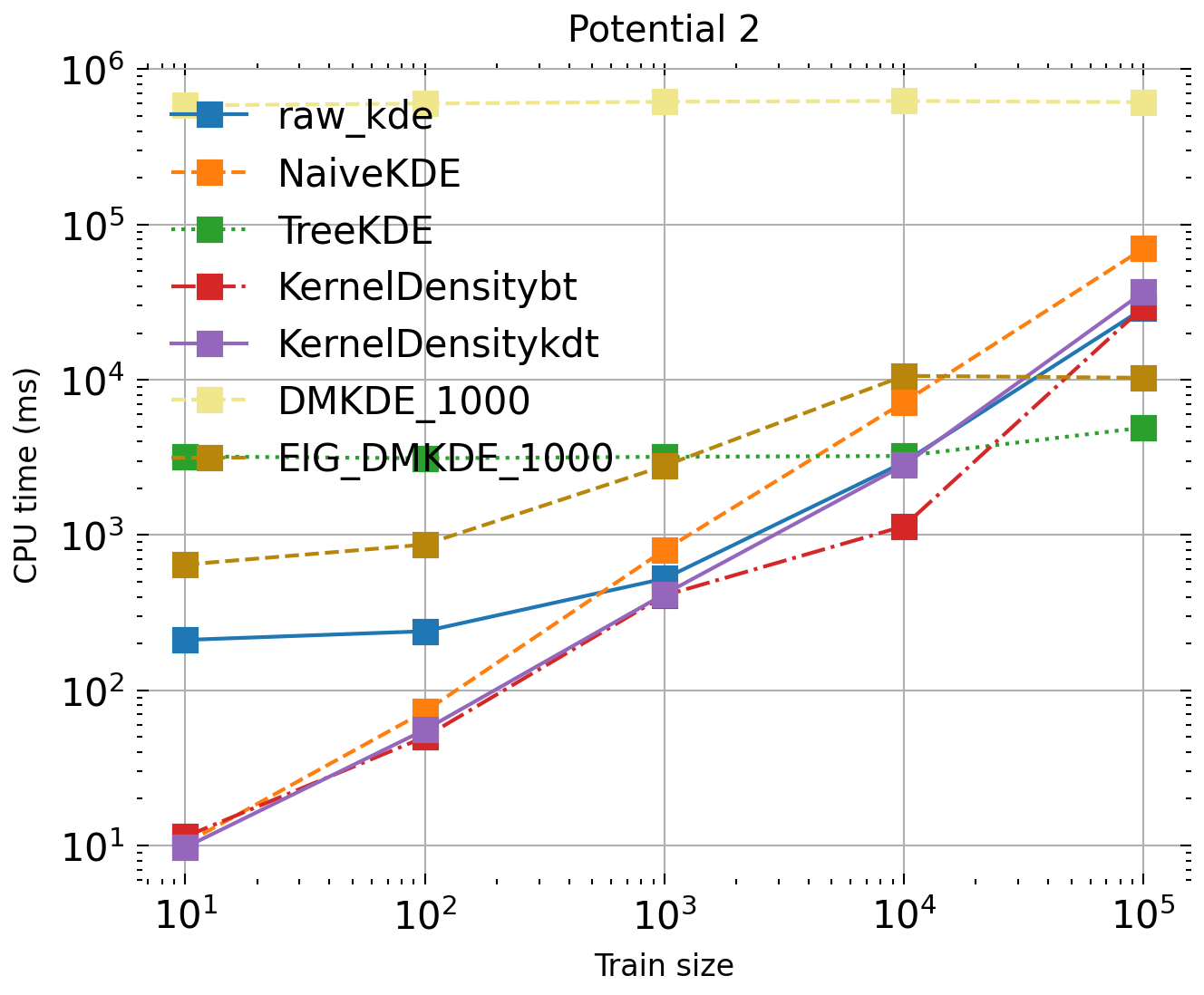}
\includegraphics[scale=0.5]{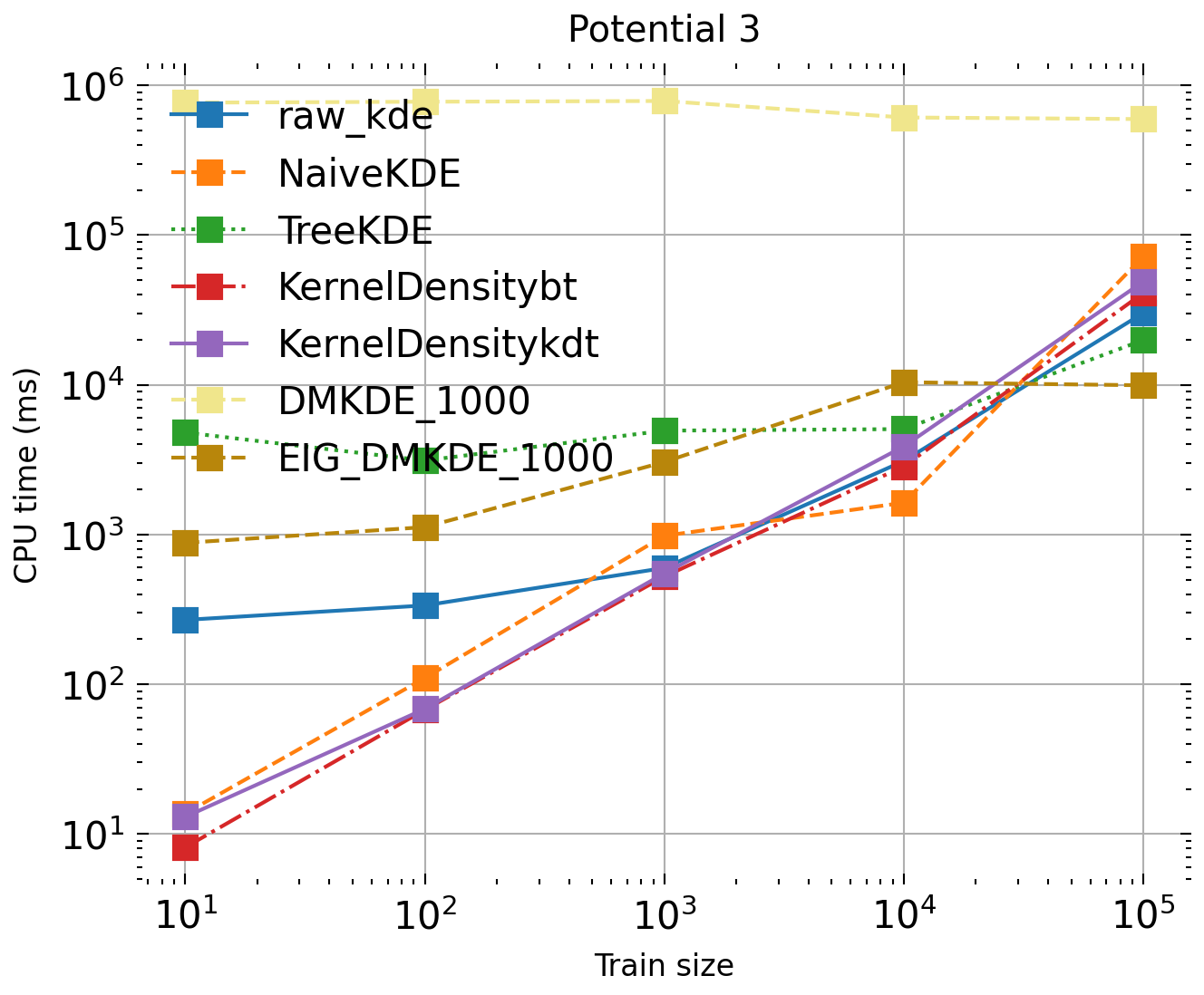}
\includegraphics[scale=0.5]{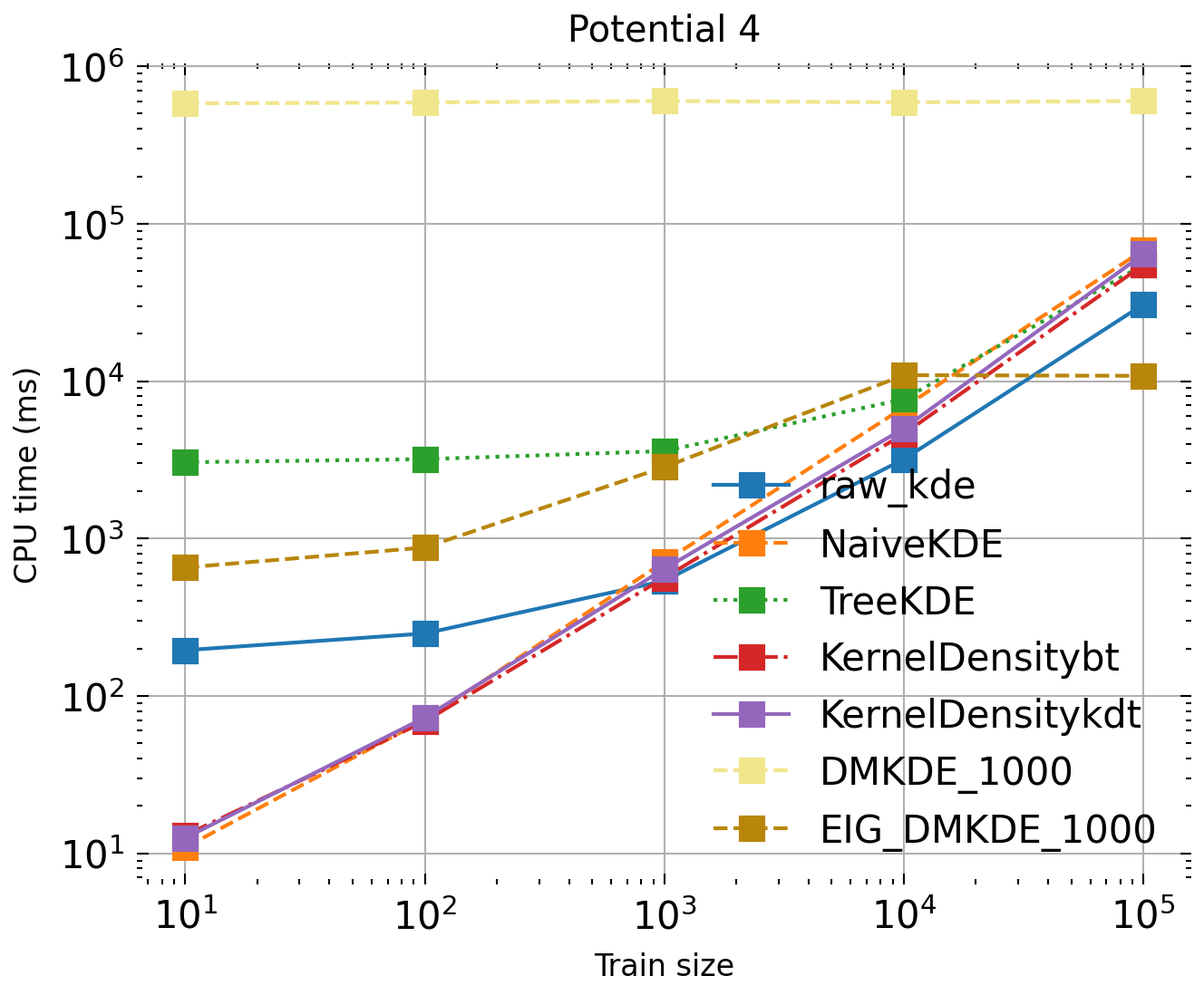}
\includegraphics[scale=0.5]{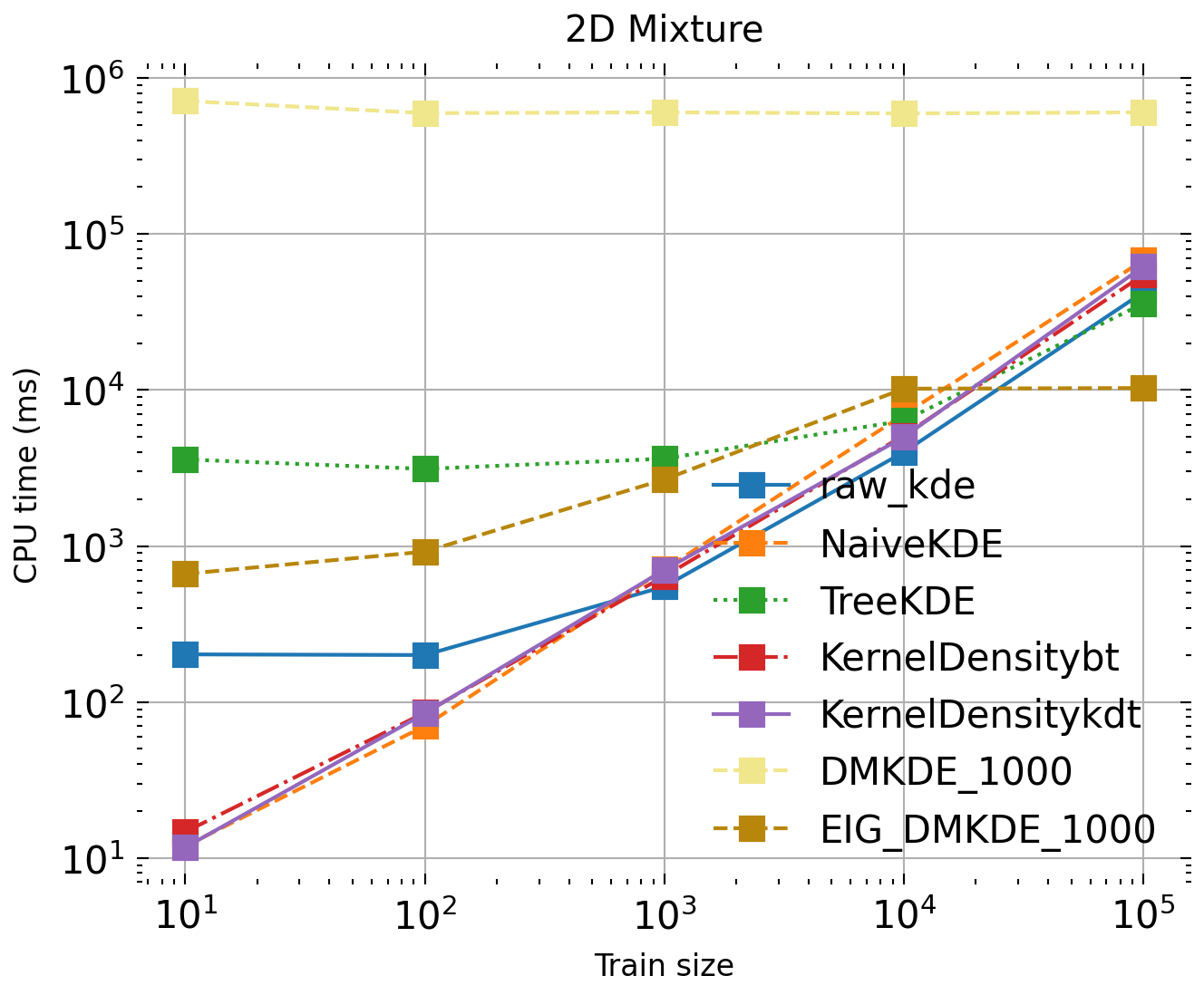}
\includegraphics[scale=0.5]{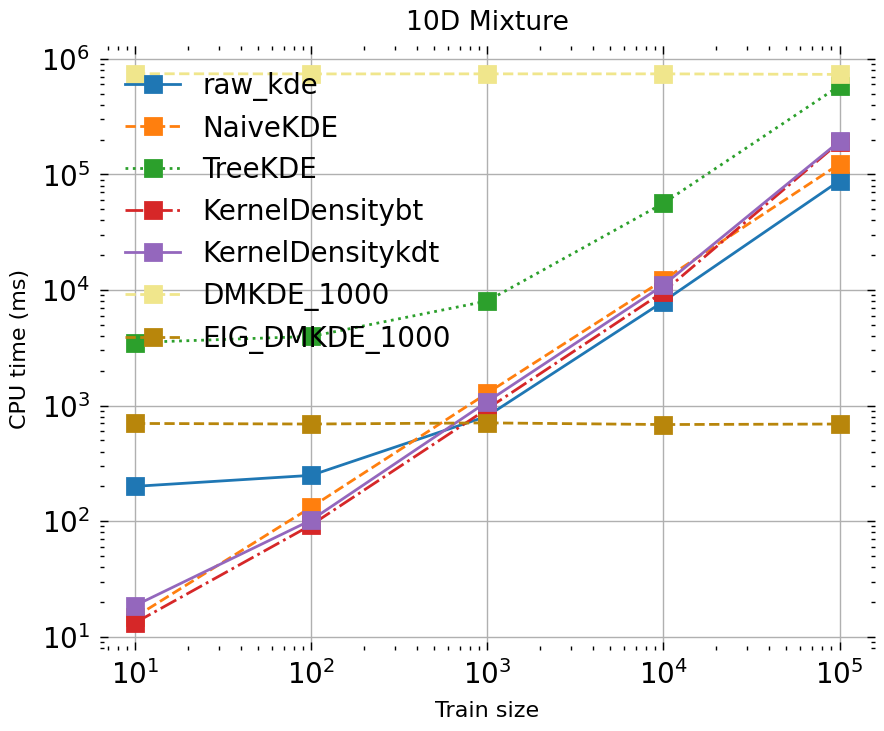}
\par\end{centering}
\caption{Comparison of the efficiency of each algorithm on six synthetic data sets. The x-axis is a logarithmic scale of $10^i$ where $i \in \{1,\cdots,5\}$. The y-axis represents the time consumed by each algorithm in milliseconds used of the central processing unit (CPU).}\label{fig:comparison_efficiency}
\end{figure*}

Table \ref{Table:mae} shows the comparison of the mean error of each algorithm on each data set against the true density. The results obtained by each approximation method are better than those of DMKDE-SGD, except for the 10-dimensional mixture data set. Table \ref{Table:time} shows the time consumed by each algorithm on each data set. The DMKDE-SGD method is at least 6 times faster than the TREE algorithm, 10 times faster than RAW, KDBTREE and KDKDTREE. And 20 times faster than NAIVE. It is worth noting that if we increase the number of training sizes, all algorithms except DMKDE-SGD will consume more time to produce a prediction. Figure \ref{fig:comparison_efficacy} shows the comparison of the efficacy measure with the mean average error (MAE) of each algorithm on six synthetic data sets. The MAE of DMKDE and DMKDE-SGD is close to other KDE approximation methods in ARC, Potential 1, Potential3, Potential 4, and 2d mixture. In Arc, however, their performance does not improve after $10^3$ training data points. In Potential 2, both DMKDE and DMKDE-SGD are better than the KDE approximation methods. On 10D Mixture, DMKDE and DMKDE-SGD outperform other KDE approximation methods. Figure \ref{fig:comparison_efficiency} shows the comparison of the efficiency measure in time taken of the central processing unit (CPU) of approximation methods of KDE. All approximation methods, except DMKDE and DMKDE-SGD, increase their prediction time when the number of points increases. However, it is observed that DMKDE does not increase linearly like the other methods. DMKDE-SGD has a larger initial footprint, but it remains constant as the number of data points increases. If we evaluate it with more than $10^5$ points, we would expect all methods to exceed the time consumed by DMKDE-SGD.

\section{Conclusion}
\label{sec:conclusions}
In this paper we systematically evaluate the performance of the method called Density Matrix Kernel Density Estimation (DMKDE). This method uses the kernel approximation given by random Fourier features and density matrices which are a fundamental tool in quantum mechanics. The efficiency and efficacy of the model was compared with three kernel density approximation methods: tree kernel density estimation method, kernel density estimation using a k-dimensional tree and kernel density using a ball tree. Systematic comparison shows that the new method is close in terms of mean error to these kernel density estimation approaches, but uses ten times less computational resources. The method can be used in domains where the size of the training data set is large (>$10^4$), where  kernel density estimation will suffer given its memory-based behavior.

\bibliography{paper}

\begin{thebibliography}{10}
\providecommand{\url}[1]{\texttt{#1}}
\providecommand{\urlprefix}{URL }
\providecommand{\doi}[1]{https://doi.org/#1}

\bibitem{Anderson2009}
Anderson, T.K.: {Kernel density estimation and K-means clustering to profile
  road accident hotspots}. Accident Analysis and Prevention  \textbf{41}(3),
  359--364 (may 2009)

\bibitem{Backurs2019}
Backurs, A., Indyk, P., Wagner, T.: Space and time efficient kernel density
  estimation in high dimensions. vol.~32 (2019)

\bibitem{bentley1975multidimensional}
Bentley, J.L.: Multidimensional binary search trees used for associative
  searching. Communications of the ACM  \textbf{18}(9),  509--517 (1975)

\bibitem{Borruso2008}
Borruso, G.: Network density estimation: a gis approach for analysing point
  patterns in a network space. Transactions in GIS  \textbf{12}(3),  377--402
  (2008)

\bibitem{devroye1985nonparametric}
Devroye, L.: Nonparametric density estimation. The L\_1 View  (1985)

\bibitem{Downs2010}
Downs, J.A.: Time-geographic density estimation for moving point objects (2010)

\bibitem{Gallego_M_Fast_Kernel_Density_2022}
Gallego~M., J.A., Osorio, J.F., Gonzalez, F.A.: {Fast Kernel Density Estimation
  with Density Matrices and Random Fourier Features Software} (7 2022).
  \doi{10.5281/zenodo.6941020}

\bibitem{gonzalez2021learning}
González, F.A., Gallego, A., Toledo-Cortés, S., Vargas-Calderón, V.:
  Learning with density matrices and random features (2021)

\bibitem{gramacki2018nonparametric}
Gramacki, A.: Nonparametric kernel density estimation and its computational
  aspects. Springer (2018)

\bibitem{GuardnacciaClaudio2021}
Guardnaccia, C., Grimaldi, M., Graziuso, G., Mancini, S.: {Crowdsourcing noise
  maps analysis by means of kernel density estimation} pp. 1691--1697 (2021)

\bibitem{Kamalov2020}
Kamalov, F.: Kernel density estimation based sampling for imbalanced class
  distribution. Information Sciences  \textbf{512},  1192--1201 (2020)

\bibitem{Kristan2011}
Kristan, M., Leonardis, A., Skočaj, D.: Multivariate online kernel density
  estimation with gaussian kernels. Pattern Recognition  \textbf{44},
  2630--2642 (10 2011)

\bibitem{Lee2012}
Lee, J., Park, M.: {An adaptive background subtraction method based on kernel
  density estimation}. Sensors (Switzerland)  \textbf{12}(9),  12279--12300
  (sep 2012)

\bibitem{Liu2020}
Liu, D., Yao, Z., Zhang, Q.: {Quantum-Classical Machine learning by Hybrid
  Tensor Networks}. Tech. rep. (2020)

\bibitem{Lv2020}
Lv, P., Yu, Y., Fan, Y., Tang, X., Tong, X.: Layer-constrained variational
  autoencoding kernel density estimation model for anomaly detection.
  Knowledge-Based Systems  \textbf{196} (5 2020)

\bibitem{josephquantum}
M., J.A.G., González, F.A.: Quantum adaptive fourier features for neural
  density estimation (2022). \doi{10.48550/ARXIV.2208.00564},
  \url{https://arxiv.org/abs/2208.00564}

\bibitem{maneewongvatana1999s}
Maneewongvatana, S., Mount, D.M.: It’s okay to be skinny, if your friends are
  fat. In: Center for geometric computing 4th annual workshop on computational
  geometry. vol.~2, pp.~1--8 (1999)

\bibitem{Nakaya2010}
Nakaya, T., Yano, K.: Visualising crime clusters in a space-time cube: An
  exploratory data-analysis approach using space-time kernel density estimation
  and scan statistics. Transactions in GIS  \textbf{14},  223--239 (6 2010)

\bibitem{omohundro1989five}
Omohundro, S.M.: Five balltree construction algorithms. International Computer
  Science Institute Berkeley (1989)

\bibitem{Papamakarios2017}
Papamakarios, G., Pavlakou, T., Murray, I.: Masked autoregressive flow for
  density estimation  (5 2017), \url{http://arxiv.org/abs/1705.07057}

\bibitem{parzen1962estimation}
Parzen, E.: On estimation of a probability density function and mode. The
  annals of mathematical statistics  \textbf{33}(3),  1065--1076 (1962)

\bibitem{Peng2019}
Peng, K., Ping, W., Song, Z., Zhao, K.: Non-autoregressive neural
  text-to-speech  (5 2019), \url{http://arxiv.org/abs/1905.08459}

\bibitem{rahimi2007rff}
Rahimi, A., Recht, B.: Random features for large-scale kernel machines. In:
  Proceedings of the 20th International Conference on Neural Information
  Processing Systems. p. 1177–1184. NIPS'07, Curran Associates Inc. (2007)

\bibitem{rezende}
Rezende, D.J., Mohamed, S.: Variational inference with normalizing flows (2015)

\bibitem{rosenblatt1956}
Rosenblatt, M.: Remarks on some nonparametric estimates of a density function.
  Ann. Math. Statist.  \textbf{27}(3),  832--837 (09 1956)

\bibitem{rudin1962fourier}
Rudin, W.: Fourier analysis on groups, vol. 121967. Wiley Online Library (1962)

\bibitem{Siminelakis2019}
Siminelakis, P., Rong, K., Bailis, P., Charikar, M., Levis, P.: Rehashing
  kernel evaluation in high dimensions. vol. 2019-June, pp. 10153--10173 (2019)

\bibitem{von1927wahrscheinlichkeitstheoretischer}
Von~Neumann, J.: Wahrscheinlichkeitstheoretischer aufbau der quantenmechanik.
  Nachrichten von der Gesellschaft der Wissenschaften zu G{\"o}ttingen,
  Mathematisch-Physikalische Klasse  \textbf{1927},  245--272 (1927)

\end{thebibliography}
\bibliographystyle{splncs04}
\end{document}